\documentclass[journal]{IEEEtran}
\usepackage{array}
\usepackage[caption=false,font=normalsize]{subfig}
\usepackage{textcomp}
\usepackage{stfloats}
\usepackage{url}
\usepackage{verbatim}
\usepackage{graphicx}
\usepackage{cite}
\usepackage{cite}
\usepackage{amsmath, amssymb, amsfonts, amsthm, mathtools}
\usepackage{algorithmic, algorithm}
\usepackage{graphicx}
\usepackage{textcomp}
\usepackage{xcolor}
\usepackage{hyperref}

\usepackage{soul} 

\usepackage{xspace} 
\newcommand{\algname}{\textsc{DT-GO}\xspace}

\newcommand{\norm}[1]{\left\lVert#1\right\rVert}
\newcommand{\normsq}[1]{\left\lVert#1\right\rVert^2}
\newcommand{\expec}[1]{\mathbb{E}\left[#1\right]}

\newcommand{\R}{\mathbb{R}}

  \DeclareMathOperator*{\diag}{diag}

\usepackage{etoolbox}
\usepackage{enumerate}
\usepackage[capitalize]{cleveref}
\AtBeginEnvironment{appendices}{\crefalias{section}{appendix}}

\crefname{figure}{Fig.}{Figs.}
\crefname{definition}{Def.}{Defs.}
\crefname{equation}{Eq.}{Eqs.}
\crefname{algorithm}{Alg.}{Algs.}
\crefname{lemma}{Lemma}{Lemmas}
\crefname{theorem}{Theorem}{Theorems}
\crefname{section}{Section}{Sections}
\crefname{proposition}{Proposition}{Propositions}

\theoremstyle{plain}
\newtheorem{theorem}{Theorem}[section]
\newtheorem*{theorem*}{Theorem}
\newtheorem{proposition}[theorem]{Proposition}
\newtheorem{lemma}[theorem]{Lemma}

\theoremstyle{definition}
\newtheorem{definition}{Definition}
\newtheorem{assumption}{Assumption}

\theoremstyle{remark}

\newenvironment{myassumption}[1]
  {\renewcommand\theassumption{#1}\begin{assumption}}
  {\end{assumption}}

\crefname{assumption}{Assumption}{Assumptions}
\Crefname{assumption}{Assumption}{Assumptions}

\usepackage{multirow, booktabs} 

\usepackage{pgfplots}
  \pgfplotsset{compat=newest}
  \usetikzlibrary{plotmarks}
  \usetikzlibrary{arrows.meta}
  \usetikzlibrary{positioning}
  \usepgfplotslibrary{patchplots}
  \usepackage{grffile}

\usepackage{pifont} 
\newcommand{\cmark}{\textcolor{green}{\ding{51}}}%
\newcommand{\xmark}{\textcolor{red}{\ding{55}}}%

\begin{document}

\title{Decentralized Optimization in Time-Varying Networks with Arbitrary Delays}

\author{Tomas Ortega, \IEEEmembership{Graduate Student Member, IEEE}, and Hamid Jafarkhani, \IEEEmembership{Fellow, IEEE} %
    \thanks{
        Authors are with the Center for Pervasive Communications \& Computing and  EECS Department, University of California, Irvine, Irvine, CA 92697 USA (e-mail: \{tomaso, hamidj\}@uci.edu). This work was supported in part by the NSF Award ECCS-2207457. An early version of this work was presented at the 2024 International Conference on Communications~\cite{dtgo}} }

{}


\maketitle

\begin{abstract}
We consider a decentralized optimization problem for networks affected by communication delays. 
Examples of such networks include collaborative machine learning, sensor networks, and multi-agent systems.
To mimic communication delays, we add virtual non-computing nodes to the network, resulting in directed graphs.
This motivates investigating decentralized optimization solutions on directed graphs.
Existing solutions assume nodes know their out-degrees, resulting in limited applicability. 
To overcome this limitation, we introduce a novel gossip-based algorithm, called \algname, that does not need to know the out-degrees.
The algorithm is applicable in general directed networks, for example networks with delays or limited acknowledgment capabilities.
We derive convergence rates for both convex and non-convex objectives, showing that our algorithm achieves the same complexity order as centralized Stochastic Gradient Descent. 
In other words, the effects of the graph topology and delays are confined to higher-order terms. 
Additionally, we extend our analysis to accommodate time-varying network topologies.
Numerical simulations are provided to support our theoretical findings.
\end{abstract}

\begin{IEEEkeywords}
	Decentralized optimization, gossip algorithms, networks with delays, collaborative machine learning, non-convex optimization.
\end{IEEEkeywords}

\section{Introduction}
\begin{figure}[htbp]
	\centering
	\tikzstyle{edge} = [very thick, -{Latex}]
\tikzstyle{vedge} = [very thick, magenta, dashdotted, -{Latex}]
\tikzstyle{non-virtualnode} = [draw, circle, fill=white, very thick]
\tikzstyle{virtualnode} = [draw, circle, magenta, fill=magenta!5, very thick, dashdotted]

\begin{tikzpicture}

    \node[non-virtualnode] (3) {\textbf{3}};
    \node[non-virtualnode] (4) [left = 1cm of 3] {\textbf{4}};
    \node[non-virtualnode] (1) [below = 1cm of 3] {\textbf{1}};
    \node[non-virtualnode] (2) [below = 1cm of 4] {\textbf{2}};
    \node[non-virtualnode] (5) [right = 1cm of 3] {\textbf{5}};

    \draw[edge] (3) -- (4);
    \draw[edge] (4) -- (2);
    \draw[edge] (2) -- (3);
    \draw[edge] (2) -- (1);
    \draw[edge] (1) -- (3);
    \draw[edge] (5) to[bend left=10] (3);
    \draw[edge] (3) to[bend left=10] (5);

    \draw[edge] (3) to[loop above, looseness=10] (3);
    \draw[edge] (4) to[loop left, looseness=10] (4);
    \draw[edge] (1) to[loop right, looseness=10] (1);
    \draw[edge] (2) to[loop left, looseness=10] (2);
    \draw[edge] (5) to[loop right, looseness=10] (5);

\end{tikzpicture}
	\caption{A directed communication graph example, with $N=5$.}
	\label{fig:example-digraph}
\end{figure}
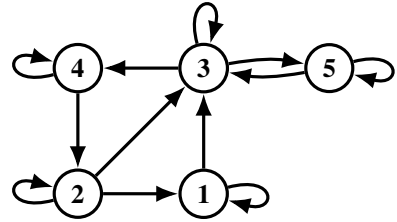

In many applications, such as decentralized estimation in sensor networks, collaborative machine learning, or decentralized coordination of multi-agent systems, the goal is to minimize a global objective function that is the average of local objective functions.
Formally, we model the network as a graph $G$, including $N$ nodes, representing sensors, agents, etc., and the corresponding edges, representing communication links -- see \cref{fig:example-digraph} for an example.
We start with a time-invariant network, where the graph edges are fixed throughout time, and extend the results to time-varying networks in \cref{sec:time-varying-analysis}.
To formalize the objective functions, we define each node's local objective function $f_n : \R^d \to \R$, as the expectation of a stochastic cost function $f_n(x) \coloneqq \mathbb{E}_{\xi_n \sim \mathcal{D}_n }[F_n(x, \xi_n)]$. 
Here, $\xi_n$ samples the local distribution $\mathcal{D}_n$, which can vary from node to node.
The nodes collaborate to find a common model $x \in \R^d$ that minimizes a global objective function $f$ of the form
\begin{equation} \label{eq:min-problem}
	f(x) \coloneqq \frac{1}{N} \sum_{n=1}^N f_n(x).
\end{equation}
Throughout the paper, we use the terms ``cost'' and ``objective'' functions interchangeably. 
\subsection{Previous work}
There exist multiple techniques to minimize~\cref{eq:min-problem}.
To the best of our knowledge, the first work studying this problem is~\cite{tsitsiklis1984problems}.
Gossip algorithms~\cite{Kempe_Dobra_Gehrke_2003,gossip_algorithms}, suggested for finding a common solution among nodes,  borrow ideas from mixing in Markov chains to allow averaging over graphs.
A combination of gradient descent with gossip steps~\cite{distributed_subgradient} has been proposed to minimize~\cref{eq:min-problem}. 
In addition, problem-specific methods such as alternating direction method of multipliers (ADMM)~\cite{Wei_Ozdaglar_2012} are used as well.
Most research efforts consider undirected communication graphs, in various settings like asynchronous communications~\cite{Srivastava_Nedic_2011}, time-varying graphs~\cite{time-varying-undirected-graphs, unified_decentralized_SGD}, or quantized communications~\cite{chocosgd,OMKL, ortegaGossip}.
The case where $G$ is directed is studied less.
In most existing algorithms, nodes \emph{must know their out-degree} to achieve consensus~\cite{decentralized_optimization_survey_nedic}.
The knowledge of the out-degree is exploited in a family of algorithms known as Push-Sum methods~\cite{weighted_gossip,distributed_subgradient}.
These have also been extended to alleviate several real-world limitations of communication networks, such as adding quantization~\cite{chocosgd}, and asynchronous communications~\cite{SGP-distributed}.

For excellent surveys on decentralized optimization, see~\cite{decentralized_optimization_survey_nedic} for synchronous methods, and~\cite{advances_in_asynchronous_optimization} for asynchronous methods.

After the publication of \cite{dtgo}, the authors were made aware of prior work in decentralized optimization that proposes algorithms which do not require out-degree knowledge.
A comparison is presented in \Cref{tab:comp}.
\begin{table*}[htbp]
\centering
\caption{Related work comparison.}
\label{tab:comp}
\begin{tabular}{|c|cccc|cc|}
\hline
\multirow{3}{*}{\textbf{Work}} &
  \multicolumn{4}{c|}{\textbf{Algorithm characteristics}} &
  \multicolumn{2}{c|}{\textbf{Analysis}} \\ \cline{2-7} 
 &
  \multicolumn{1}{c|}{\textbf{Only in-degree}} &
  \multicolumn{1}{c|}{\multirow{2}{*}{\textbf{Time-Varying}}} &
  \multicolumn{1}{c|}{\textbf{No warm-up}} &
  \textbf{No extra} &
  \multicolumn{1}{c|}{\multirow{2}{*}{\textbf{Non-convex objective}}} &
  \textbf{Allows unbounded} \\
 &
  \multicolumn{1}{c|}{\textbf{knowledge}} &
  \multicolumn{1}{c|}{} &
  \multicolumn{1}{c|}{\textbf{phase}} &
  \textbf{variables communicated} &
  \multicolumn{1}{c|}{} &
  \textbf{gradients} \\ \hline
\cite{mai2016distributed} &
  \multicolumn{1}{c|}{\cmark} &
  \multicolumn{1}{c|}{\xmark} &
  \multicolumn{1}{c|}{\cmark} &
  \xmark &
  \multicolumn{1}{c|}{\xmark} &
  \xmark \\ \hline
\cite{xi2018linear} &
  \multicolumn{1}{c|}{\cmark} &
  \multicolumn{1}{c|}{\xmark} &
  \multicolumn{1}{c|}{\cmark} &
  \xmark &
  \multicolumn{1}{c|}{\xmark} &
  \cmark \\ \hline
This work &
  \multicolumn{1}{c|}{\cmark} &
  \multicolumn{1}{c|}{\cmark} &
  \multicolumn{1}{c|}{\xmark} &
  \cmark &
  \multicolumn{1}{c|}{\cmark} &
  \cmark \\ \hline
\end{tabular}
\end{table*}

\subsection{Contributions}
Existing algorithms for directed decentralized optimization assume that nodes in the network  know their out-degree.
In this paper, we propose a gossip-based decentralized optimization solution for which nodes \emph{do not} need to know their out-degree to achieve convergence.
Such a property is desired in many scenarios which arise naturally.
For example, consider networks with directed links established without a handshake, i.e, transmitters send messages and the receivers do not need to acknowledge their reception.
Such a setting is likely in heterogeneous network scenarios where some nodes have smaller transmit powers, making bidirectional communication impossible for some links.
Networks with broadcast capabilities are another example as broadcasting nodes do not know who receive their messages. 
We present an algorithm, called Delay Tolerant Gossiped Optimization (\algname), that allows decentralized optimization in such networks.

The main contributions of our work can be summarized as follows.
We propose a decentralized stochastic optimization algorithm that
\begin{itemize}
    \item works on time-varying directed communication networks, where nodes do not know how many listeners they have, and is robust in the presence of communication delays;
    \item has convergence guarantees for $L$-smooth convex and non-convex objectives, where topology and delays are shown to affect only the higher order terms;
    \item does not need multiple rounds of gossip per each round of local optimization or bounded gradient assumptions to guarantee convergence;
    \item generalizes decentralized SGD, as for the case of bidirectional communication graphs, it recovers decentralized SGD both in theory and practice;
    \item is shown to converge, as analytically proven, on a set of logistic regression problems, with directed time-varying networks suffering from communication delays.
\end{itemize}

\subsection{Organization}
The rest of the paper is organized as follows.
\Cref{sec:setup-and-proposed-algorithm} formally sets up our approach to minimize \cref{eq:min-problem}.
First, the problem is divided into optimization and decentralized averaging in \cref{sec:setup}.
Then, we illustrate the challenges with existing decentralized averaging schemes in \cref{sec:decentralized-averaging}.
We provide a numerical toy example to illustrate our theoretical insights.
This gives way to the algorithm design, which is described in \cref{sec:algorithm-design}.
The framework to extend our setup to incorporate delays is illustrated in \cref{sec:incorporating-delays}.
We then theoretically analyze our algorithm for time-invariant communication graphs in \cref{sec:analysis}, first for cases without delays, and afterwards with delays.
We extend our analysis to time-varying graphs in \cref{sec:time-varying-analysis} and provide convergence guarantees for convex and non-convex objectives.
Finally, we present a set of experimental results for logistic regression problems in \cref{sec:experimental-results} and provide the paper conclusions in \cref{sec:conclusions}.

\subsection{Notation}
We denote a vector corresponding to Agent $n$ and Round $k$ as $x_n^{[k]}$.
The matrix whose $N$ rows are the vectors $x_1^{[k]}, \ldots, x_N^{[k]}$ is denoted $X^{[k]}$. 
Note that subscripts identify agents and superscripts between brackets identify rounds.
The expectation operator is denoted $\mathbb{E}[\cdot]$.
We denote the 2-norm and Frobenius norm of $x$ as $\norm{x}_2$ and $\norm{x}_F$, respectively.

\section{Proposed algorithm} \label{sec:setup-and-proposed-algorithm}
\subsection{Problem setup} \label{sec:setup}
We consider a \emph{directed} graph $G$ including $N$ nodes that share their information over the network modeled by the graph. 
We assume the nodes can \emph{only} know their \emph{in-degree} and communicate with their directed neighbors since there is no handshake mechanism to know the out-degrees. 

To minimize~\cref{eq:min-problem}, we iterative over the following two phases: (i) local optimization, and (ii) consensus.
At each local optimization phase, nodes optimize their local models using their local data.
This makes local models drift from the average, and the problem of making them drift back to the average is known as decentralized averaging.
To solve decentralized averaging, we introduce a consensus phase, where nodes communicate among themselves to converge to the average of their models.

\subsection{Decentralized averaging}\label{sec:decentralized-averaging}
In this section, we introduce a decentralized averaging algorithm that allows nodes' states to converge to the average of all states without using their out-degrees.
Let us denote the initial model, or initial state, of Node $n$ as $x_n^{[0]}$, which is a vector in $\R^d$.
The state of Node $n$ at iteration $k$ is denoted by $x_n^{[k]}$.
At each iteration, nodes broadcast their models and collect the neighbors' models.
Then, the nodes perform a \emph{gossip} step, i.e., calculate the weighted average over the received models.
Specifically, Node $n$ weighs information received from Node $m$ with weight $W_{nm}$.
A natural choice for these weights is the inverse of the in-degree of each node.
For example, for the graph in \cref{fig:example-digraph}, the matrix $W$ can be
\begin{equation} \label{eq:example-gossip-matrix}
    W = \begin{pmatrix}
        \tfrac{1}{2} & \tfrac{1}{2} & 0 & 0 & 0 \\
        0 & \tfrac{1}{2} & 0 & \tfrac{1}{2} & 0 \\
        \tfrac{1}{4} & \tfrac{1}{4} & \tfrac{1}{4} & 0 & \tfrac{1}{4} \\
        0 & 0 & \tfrac{1}{2} & \tfrac{1}{2} & 0 \\
        0 & 0 & \tfrac{1}{2} & 0 & \tfrac{1}{2}
        \end{pmatrix}.
\end{equation}
Other choices also work, as long as they satisfy our definition of a gossip matrix.
\begin{definition}[Gossip matrix]\label{def:gossip-matrix}
	Given a strongly connected directed graph $G$, its gossip matrix $W$ is an $N\times N$ real matrix whose entries satisfy:
	\begin{enumerate}[(i)]
		\item $W$ is row-stochastic: the sum of the entries in every row is one, i.e., $\sum_{m=1}^N W_{nm} = 1$ for all $n$.
		\item Every entry $W_{nm}$ is non-negative, and zero only if there is no directed edge from Node $m$ to Node $n$ in $G$.
		\item Diagonal entries are positive, i.e., each node has a self-loop.
	\end{enumerate}
\end{definition}
Formally, each gossip iteration is 
\begin{equation}
    x_n^{[k+1]} = \sum_{m=1}^N W_{nm} x_m^{[k]},
\end{equation}
which can be written in a matrix form as
\begin{equation} \label{eq:gossip-iteration}
    X^{[k+1]} = W X^{[k]}.
\end{equation}
Here, $X^{[k]}$ is the $N \times d$ real matrix whose rows are the node states.
Notice from Item (ii) in \cref{def:gossip-matrix} that $W_{nm}$ is only positive when Node $n$ can actually receive model $x_m^{[k]}$.

\begin{figure}[htbp]
	\centering
    \includegraphics[width=\columnwidth]{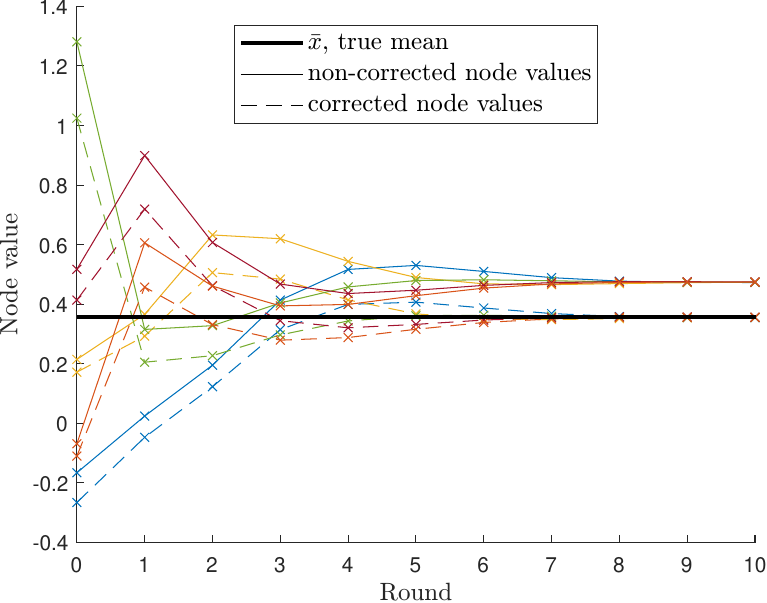} 
	\caption{Plot of corrected and non-corrected node values at different rounds. Initial node values are chosen at random from a normal $\mathcal{N}(0,5)$. The gossip weights are the inverse of the in-degrees, where $G$ is shown in \cref{fig:example-digraph}.}
	\label{fig:corrected-values}
\end{figure}

The goal is to generate a sequence of models at every node that converges to the average, i.e, 
\begin{equation}
    \lim_{k \to \infty} x_n^{[k]} \to \bar x = \frac{1}{N} \sum_{n=1}^N x_n^{[0]},
\end{equation}
for all nodes $n$.
However, the local models \emph{do not} necessarily converge to the average if the gossip matrix is row-stochastic.
\cref{fig:corrected-values} shows an example in which the nodes using a regular gossip algorithm, marked by the ``non-corrected node values'', converge but not to $\bar x$. 
It can be shown that instead the nodes converge to a weighted average 
\begin{equation}\label{eq:weighted-average}
	\tilde x \coloneqq \sum_{n=1}^N \pi_n x_n^{[0]},
\end{equation}
where $\pi_n$ are non-negative weights that add up to one~\cite[Lemma 5]{nedic_consensus}.
We later prove in \cref{sec:analysis} that $\pi_1, \ldots, \pi_N$ are positive.
In this section, first, we present an algorithm that ``corrects'' the values such that the local models converge to the average.
Then, we show how to handle links with communication delays using our algorithm.

\subsection{Algorithm design}\label{sec:algorithm-design}
We present \algname, which corrects the weighted average described in \cref{eq:weighted-average} and allows the consensus phase to converge to the true average.
As we introduced in \cref{sec:setup}, we later interchange consensus and local optimization phases to minimize the cost function in \cref{eq:min-problem}.

The key idea is described in the following Lemma.
\begin{lemma}
	Consider a strongly connected directed graph $G$ with an associated gossip matrix $W$ as in \cref{def:gossip-matrix}.
	If every node $n$ multiplies its initial state $x_n^{[0]}$ by a factor $d_n \coloneqq \frac{1}{N \pi_n}$, then the gossip iterations $$x_n^{[k+1]} = \sum_{m=1}^N W_{nm} x_m^{[k]}$$ converge to the true mean, i.e.,
	\begin{equation}
		\sum_{n=1}^N \pi_n d_n x_n^{[0]} = \sum_{n=1}^N \frac{1}{N} x_n^{[0]} = \bar x.
	\end{equation}
\end{lemma}
The proof of this lemma follows immediately from \cref{eq:weighted-average} by inserting the re-weighted initial states.

We illustrate this lemma with a toy example in \cref{fig:corrected-values}.
Re-weighting the initial state of Node $n$ with $d_n$ for all $n$ in \cref{fig:example-digraph} results in curves labeled with corrected node values. These curves converge to the true average of the initial state.
In contrast, curves labeled non-corrected node values, related to the original gossip algorithm, converge to a weighted average $\tilde x$.

Thus, it is paramount to obtain the correction weights $d_n \coloneqq \frac{1}{N \pi_n}$ to converge to the real average. 
Let us start with assuming that $N$ is known to all nodes. 
We will relax this assumption later. 
To compute $d_n$, nodes must obtain $\pi_n$ first. 
We observe that if every node starts with an initial state $x_n^{[0]} = e_n$, a vector of length $N$ that includes all zeros, except a one in coordinate $n$, the gossip iterations converge to
\begin{equation}\label{eq:obtaining-pin}
	\tilde x = \sum_{n=1}^N \pi_n e_n = \pi_1 \begin{pmatrix}
	    1 \\
        0 \\
        \vdots \\
        0
	\end{pmatrix} + \cdots + \pi_N \begin{pmatrix}
	    0 \\
        \vdots \\
        0 \\
        1
	\end{pmatrix} = \begin{pmatrix}
	    \pi_1 \\
        \pi_2 \\
        \vdots \\
        \pi_N
	\end{pmatrix},
\end{equation}
that is, a vector with $\pi_n$ in the $n$-th coordinate.
Therefore, each node can compute $d_n$ when convergence is achieved by simply retrieving $\pi_n$ from the $n$-th coordinate and dividing by $N$.
Leveraging this observation, we propose \cref{alg:decentralized_opt_alg}.
\begin{algorithm}[b!] 
	\caption{\algname at Node $n$}\label{alg:decentralized_opt_alg}
	\begin{algorithmic}[1]
		\STATE Generate local id number $id_n$. \\ \COMMENT {\textsc{Warm-Up Period}}
		\STATE {Initialize dictionary $dict \gets \{id_n:1\}$.}
		\FOR{$K_{\text{warm-up}}$ rounds}
		\STATE {Broadcast $dict$ and receive neighbors' dictionaries.}
		\STATE {$dict \gets$ weighted average of available dictionaries.}
		\ENDFOR{}
		\STATE{From $dict$, obtain $N$ and $\pi_n$, as in \cref{eq:obtaining-pin}. \\ \COMMENT {\textsc{Minimizing $f$ Period}}}
		\STATE {Initialize $x_n^{[0]}$.}
		\FOR{$k$ in $0, \ldots, K-1$}
		\STATE {Initialize auxiliary variables $z_n, y$. \\ \COMMENT {Optimization phase}}
		\STATE {SGD step: $y \gets x_n^{[k]} - \eta \nabla F_n(x_n^{[k]}, \xi_n)$. \\ \COMMENT {Consensus phase}}
		\STATE {Adjust update: $z_n \gets x_n^{[k]} + \frac{1}{N\pi_n} (y - x_n^{[k]})$.}
		\STATE {Broadcast $z_n$ and receive neighbor states.}
		\STATE {Average states: $z_n \gets \sum_{m=1}^N W_{nm} z_m$.}
		\STATE {$x_n^{[k+1]} \gets z_n$.}
		\ENDFOR{}
	\end{algorithmic}
\end{algorithm}
Note that we do not use vectors $e_n$, because nodes do not know the network size $N$ a priori.
This challenge can be solved elegantly by using dictionaries, as done in \cref{alg:decentralized_opt_alg}.
At round zero, every node starts with a dictionary initialized with the key that corresponds to their identification and the value one.
Namely, every node $n$ initializes its dictionary by $dict = \{id_n : 1\}$.
Note that this serves as a proxy for $e_n$, where the zero entries are conveniently not in the dictionary.
Next, the dictionaries are broadcasted to the neighbors.
Elements that do not appear in neighboring dictionaries are simply treated as zero, and the algorithm remains unchanged.
The decentralized averaging is run for a certain number of rounds, denoted the warm-up period, until convergence.
Once the warm-up period is over, nodes obtain $N$ from the dictionary size and compute the correction weights $d_n$.

\subsection{Incorporating Delays}\label{sec:incorporating-delays}
The framework proposed for \algname is designed to easily accommodate links with delays, i.e., links where information takes more than one round to arrive at the receiver.
To incorporate the delays, similar to~\cite{nedic_consensus, tsitsiklis1984problems}, we introduce the notion of virtual nodes, or non-computing nodes, that serve as a relay of the message for a round.
If Node $n$ sends messages to Node $m$ with a delay of $\ell$ rounds, we simply add $\ell$ nodes to the graph $G$, each with objective function 0 to avoid modifying the objective function in~\cref{eq:min-problem}.
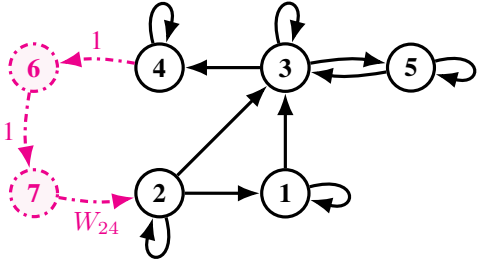
\begin{figure}[htbp]
	\centering
	\tikzstyle{edge} = [very thick, -{Latex}]
\tikzstyle{vedge} = [very thick, magenta, dashdotted, -{Latex}]
\tikzstyle{non-virtualnode} = [draw, circle, fill=white, very thick]
\tikzstyle{virtualnode} = [draw, circle, magenta, fill=magenta!5, very thick, dashdotted]

\begin{tikzpicture}

    \node[non-virtualnode] (3) {\textbf{3}};
    \node[non-virtualnode] (4) [left = 1cm of 3] {\textbf{4}};
    \node[non-virtualnode] (1) [below = 1cm of 3] {\textbf{1}};
    \node[non-virtualnode] (2) [below = 1cm of 4] {\textbf{2}};
    \node[non-virtualnode] (5) [right = 1cm of 3] {\textbf{5}};
    \node[virtualnode] (6) [left = 1cm of 4] {\textbf{6}};
    \node[virtualnode] (7) [left = 1cm of 2] {\textbf{7}};

    \draw[vedge] (4) to[bend right=10] node[above]{1} (6);
    \draw[vedge] (6) to[bend right=10] node[left]{1} (7);
    \draw[vedge] (7) to[bend right=10] node[below]{$W_{24}$} (2);
    \draw[edge] (3) -- (4);
    \draw[edge] (2) -- (3);
    \draw[edge] (2) -- (1);
    \draw[edge] (1) -- (3);
    \draw[edge] (5) to[bend left=10] (3);
    \draw[edge] (3) to[bend left=10] (5);

    \draw[edge] (3) to[loop above, looseness=10] (3);
    \draw[edge] (4) to[loop above, looseness=10] (4);
    \draw[edge] (1) to[loop right, looseness=10] (1);
    \draw[edge] (2) to[loop below, looseness=10] (2);
    \draw[edge] (5) to[loop right, looseness=10] (5);

\end{tikzpicture}
	\caption{Example of a graph with delayed links. We have added a delay of 2 rounds to the edge between Node 4 and Node 2 of the graph in \cref{fig:example-digraph}.}\label{fig:delayed-graph}
\end{figure}
For example, in \cref{fig:delayed-graph}, Node 4 sends messages to Node 2 with a delay of 2 rounds.
Therefore, to accommodate the delay, we add two nodes, 6 and 7, to the graph and modify the connections as follows: Node 4 sends messages to Node 6, which sends messages to Node 7, which finally sends messages to Node 2.
The weight of the old edge from Node 4 to 2, $W_{24}$, is now assigned to the edge from Node 7 to 2 and the weights of the other new edges are set to 1.
If we set the edge weights to be the inverse of the in-degrees as in \cref{eq:example-gossip-matrix}, the extended weight matrix $\mathcal{W}$ is
\begin{equation} \label{eq:example-gossip-matrix-extended}
    \mathcal{W} = \begin{pmatrix}
        \tfrac{1}{2} & \tfrac{1}{2} & 0 & 0 & 0 & 0 & 0 \\
        0 & \tfrac{1}{2} & 0 & 0 & 0 & 0 & \mathbf{\tfrac{1}{2}} \\
        \tfrac{1}{4} & \tfrac{1}{4} & \tfrac{1}{4} & 0 & \tfrac{1}{4} & 0 & 0 \\
        0 & 0 & \tfrac{1}{2} & \tfrac{1}{2} & 0 & 0 & 0 \\
        0 & 0 & \tfrac{1}{2} & 0 & \tfrac{1}{2} & 0 & 0 \\
        0 & 0 & 0 & \mathbf{1} & 0 & 0 & 0 \\
        0 & 0 & 0 & 0 & 0 & \mathbf{1} & 0
        \end{pmatrix},
\end{equation}
where we have highlighted the terms that are involved with the virtual nodes, as shown in \cref{fig:delayed-graph}.
Crucially, this procedure ensures that the nodes' input weights \emph{still add up to one} and $\mathcal{W}$ is row-stochastic.
Also, the resulting extended graph is \emph{still strongly connected}.

\section{Time-invariant analysis} \label{sec:analysis}
In this section, first, we discuss the case without delays, as it gives a stronger and more illustrative bound.
Then, we present the case with delays.
\subsection{Case without delays} \label{sec:case-without-delays}
We introduce a proposition to show that our algorithm converges on strongly connected directed graphs.
\begin{proposition}\label{prop:gossip-properties}
	Given a gossip matrix $W$ satisfying \cref{def:gossip-matrix}, we can ensure that
	\begin{enumerate}[(i)]
		\item The limit of the gossip matrix powers $\lim_{k \to\infty} W^k \coloneqq W^\infty$ exists, that is, our gossip algorithm converges to a stationary solution.
		\item The matrix $W^\infty$ is row-stochastic and its rows are all identical, with positive entries $\pi_1, \ldots, \pi_N$ that add up to one. Namely,
        \begin{equation*}
            W^\infty = \begin{pmatrix}
                \pi_1 & \pi_2 & \ldots & \pi_N \\
                \pi_1 & \pi_2 & \ldots & \pi_N \\
                \vdots & \vdots & \ddots & \vdots \\
                \pi_1 & \pi_2 & \ldots & \pi_N
            \end{pmatrix}.
        \end{equation*}
		\item The squared Frobenius norm of the global state converges to the stationary solution at a geometric rate, i.e., there exist positive constants $C$ and $\rho$, with $\rho < 1$, such that for all $k>0$, $\norm{W^k - W^\infty}_2^2 \leq C \rho^k$.
	\end{enumerate}
\end{proposition}
\begin{IEEEproof}
	Since $W$ is primitive, Statements (i) and (ii) are immediate consequences of applying the Perron-Frobenius theorem~\cite{Meyer_Stewart_2023}.
	Furthermore, $\pi_1, \ldots, \pi_N$ are the coordinates of the left Perron eigenvector of $W$. 
	Following the ideas in~\cite[Fact 3]{convergence_rate_markov}, Statement (iii) has a straightforward proof via eigendecomposition.
	In addition, we have $\rho = |\lambda_2|^2$, the second largest eigenvalue of $W$ in absolute value squared.
\end{IEEEproof}

\subsection{Case with delays}
Let us now assume that we have a network with arbitrary delays.
The number of real nodes is $N$. 
We add virtual nodes to obtain a network of size $\mathcal{N}$.
Similarly, the original gossip matrix $W$ is extended as previously described to a gossip matrix $\mathcal{W}$.
\begin{proposition}\label{prop:gossip-properties-with-delays}
	Given a gossip matrix $W$ satisfying \cref{def:gossip-matrix} and its extended version with delays $\mathcal{W}$, we can ensure that
	\begin{enumerate}[(i)]
		\item The limit of the matrix powers $\lim_{k \to \infty} \mathcal{W}^k \coloneqq \mathcal{W}^\infty$ exists, that is, our gossip algorithm with delays converges to a stationary solution.
		\item The matrix $\mathcal{W}^\infty$ is row-stochastic and its rows are all identical, with non-negative entries $\pi_{1}, \ldots, \pi_{\mathcal{N}}$ that add up to one.
		\item The squared Frobenius norm of the global state converges to the stationary solution at a geometric rate, i.e., there exist constants $C$ and $\rho$, with $\rho < 1$, such that for all $k>0$, $\norm{\mathcal{W}^k - \mathcal{W}^\infty}_2^2 \leq C \rho^k$.
		\item The weights that correspond to non-virtual nodes are all positive, i.e., $\pi_{1}, \ldots, \pi_{N} > 0$.
	\end{enumerate}
\end{proposition}
\begin{IEEEproof}
	From~\cite[Lemma 5]{nedic_consensus}, Facts (i) and (ii) follow immediately.
	For Fact (iii),~\cite[Lemma 5]{nedic_consensus} ensures that
	\begin{equation}
		\norm{\mathcal{W}^k - \mathcal{W}^\infty}_F \leq 2 \frac{1 + \eta^{-B_2}}{1 - \eta^{B_2}}(1 - \eta^{B_2})^{\frac{k}{B_2}},
	\end{equation}
	where $B_2 \coloneqq N - 1 + NB_1$, $B_1$ is the maximum number of delays between two nodes in $G$, and $\eta \in (0,1)$ is a positive lower bound for all the non-zero entries of $\mathcal{W}$.
	Squaring the expression and knowing that the Frobenius norm upper-bounds the 2-norm proves the statement.
	The proof of (iv) follows from~\cite[Lemma 5.2.1]{tsitsiklis1984problems}.
\end{IEEEproof}
\cref{prop:gossip-properties-with-delays} guarantees convergence for graphs with delays, albeit it often provides a looser bound for delay-less graphs.
Note that in what follows, we do not use the extended weight matrix notation $\mathcal{W}$ or the extended node quantity $\mathcal{N}$ in the derivations, but the statements hold for those cases as well.
We provide extra explanations in the crucial steps.

A corollary of \cref{prop:gossip-properties,prop:gossip-properties-with-delays} is that after a number $\tau$ of gossip rounds, such that $C \rho^{\tau} < 1$, the quantity $\normsq{W^\tau - W^\infty}$ is less than $1$, as needed for our convergence results.
In other words, if \cref{def:gossip-matrix} is satisfied, our gossip phase converges to a weighted average of the initial states.
Moreover, the weighted average of states \emph{is preserved} through iterations.
This can be verified easily in matrix notation by denoting $\tilde x^{[k]}$ as the weighted average of node models at Round $k$,
\begin{equation}
    \tilde x^{[k]} \coloneqq \sum_{n=1}^N \pi_n x_n^{[k]},
\end{equation}
and the corresponding extension to matrix notation,
\begin{equation} \label{eq:weighted-average-def}
    \tilde X^{[k]} \coloneqq W^\infty X^{[k]}.
\end{equation}
Thus, by the definition of our gossip iterations from \cref{eq:gossip-iteration}, for all positive $k$,
\begin{equation} \label{eq:weighted-average-is-preserved}
    \tilde X^{[k]} = W^\infty W X^{[k-1]} = W^\infty X^{[k-1]},
\end{equation}
and, by induction, the weighted average of the initial states is preserved.

A second remark on \cref{prop:gossip-properties,prop:gossip-properties-with-delays} is that if $W$ is doubly stochastic and we do not add delays, then $C=1$ and $\tau=1$.
In addition, $\lim_{k \to \infty}W^k$ converges to an all-one matrix divided by $N$ and our algorithm is equivalent to Decentralized SGD (DSGD)~\cite{distributed_subgradient}.
Furthermore, if $G$ is a complete graph and we set edge weights to $1/N$, then the gossip matrix is the averaging matrix and both \algname and DSGD are equivalent to centralized SGD.

\section{Time-varying analysis} \label{sec:time-varying-analysis}
We now extend \cref{alg:decentralized_opt_alg} to the case of time-varying communication networks.
Note that the number of nodes remains the same, but the existence of edges at different rounds may change.
We denote the communication graph at Round $k$ as $G^{[k]}$ and its associated weight matrix as $W^{[k]}$ (which may not satisfy \cref{def:gossip-matrix}, as is discussed later).
This allows us to model several scenarios of interest, such as networks where links are available only with a certain probability, or networks that only communicate every several rounds (equivalent to multiple local SGD steps).

Assuming that the warm-up phase is complete, we redefine $X^{[k]}$ to include the corrected gradient step. Then, the combination of Steps 10-15 in 
\cref{alg:decentralized_opt_alg} results in the following matrix equation:
\begin{equation}\label{eq:alg-iterations}
	X^{[k+1]} = W^{[k]}\left(X^{[k]} - \eta D \partial F\left(X^{[k]}\right) \right),
\end{equation}
where $\partial F\left(X^{[k]}\right)$ is an $N \times d$ real matrix that represents the stochastic gradients at Round $k$ and $D \coloneqq \diag(\tfrac{1}{N\pi_1}, \ldots, \tfrac{1}{N\pi_N})$ is the diagonal correction matrix.
If there are virtual nodes, the corresponding entries in the diagonal matrix $D$ are set to $1$, ensuring $D$ is well defined.
This matrix notation allows us to adopt the following standard assumptions~\cite{unified_decentralized_SGD,sgd-gower}.

\begin{myassumption}{1}\label{a:gossip}
	Given a set of (possibly random) graphs $G^{[0]}, G^{[1]}, \ldots$ and associated $N\times N$ real weight matrices $W^{[0]}, W^{[1]}, \ldots$, where an entry is zero if and only if the corresponding edge does not exist, we assume:
	\begin{enumerate}[(i)]
		\item The limit $\lim_{k\to\infty} \expec{\prod_{s=0}^k W^{[s]}} \coloneqq W^\infty$ exists and satisfies $W^{[k]} W^\infty = W^\infty$ and $W^\infty W^{[k]} = W^\infty$ for all $k$.
		\item The matrix $W^\infty$ is row-stochastic and each of its rows contains identical non-negative entries denoted $\pi_1, \ldots, \pi_N$ that add up to one. These entries are positive for non-virtual nodes.
		\item There exists a positive constant $p \leq 1$ and integer $\tau \geq 1$ such that for all $k \geq \tau$,
		      \begin{equation}
			      \mathbb{E} \normsq{W^\infty - \prod_{\ell=0}^{\tau-1} W^{[k+\ell]}}_2 \leq 1 - p.
		      \end{equation}
		\item There exists a positive constant $\beta$ such that for all $k$ and all non-negative $\ell < \tau$,  $$\mathbb{E} \normsq{\left(\prod_{m=\ell}^{\tau-1} W^{[k+m]} D\right) - J}_2 \leq \beta^2,$$ where $D$ is the diagonal correction matrix and $J$ is the average matrix with all entries equal to $1/N$.
	\end{enumerate}
\end{myassumption}

Note that for the time-invariant case, \cref{a:gossip} is satisfied by gossip matrices satisfying \cref{def:gossip-matrix}, as shown in \cref{prop:gossip-properties,prop:gossip-properties-with-delays} for the cases without and with delays, respectively.
However, \cref{a:gossip} covers many more scenarios, for example, $G^{[k]}$ does not necessarily have to be connected at each iteration, and $W^{[k]}$ need not be row-stochastic or satisfy \cref{def:gossip-matrix}, as long as the assumption holds.

\cref{a:gossip} is an extension of \cref{prop:gossip-properties-with-delays}, simply avoiding a pathological unbounded sum of norms when defining $\beta$ in~(iv).
Note that $\beta$ is independent of $k$, and depends only on the graph topology and chosen weights.

\Cref{a:gossip} also accommodates multiple common scenarios.
For the sake of brevity, we only present a few examples.
The first is the \emph{standard bidirectional communications case}.
If $G^{[k]}$ is a bidirectional connected communication network, we can design a doubly stochastic $W^{[k]}$.
This implies that $\pi_1, \ldots, \pi_N$ are all $1/N$ and $D$ is simply the identity matrix. 
Therefore, $\tau=1$ and $\beta=1$.
Another common scenario is the \emph{multiple local steps setting}.
To recover this setting, we now consider that $G^{[k]}$ is a fixed strongly connected directed graph every $R$ rounds and is an empty graph (only self-loops exist) for every other round.
For the empty graph rounds, the gossip matrix is the identity.
Following the same reasoning from \cref{prop:gossip-properties}, one can see that this setting satisfies \cref{a:gossip}.

Let us now define the notion of $L$-smoothness, as we use it extensively for the analysis.
\begin{definition}
	We call a function $L$-smooth if it is continuously differentiable and its gradient is Lipschitz continuous with Lipschitz constant $L$. That is, for all $x,y \in \R^d$:
	\[\norm{\nabla f(x) - \nabla f(y)} \leq L \norm{x - y}.\]
\end{definition}

\subsection{Convex case}
For the convex case, we assume the stochastic gradients are $L$-Lipschitz and that the noise and heterogeneity of our cost functions is bounded at the optimum.
The precise assumptions are as follows.
\begin{myassumption}{2a}\label{a:lsmooth-stoch}
	Each function $F_n(x, \xi_n) $ is $L$-smooth with respect to $x$, that is, for all $x, y \in \R^d$:
	\begin{equation}
		\norm{\nabla F_n(x, \xi_n) - \nabla F_n(y,\xi_n) } \leq L \norm{x - y}. \label{eq:F-smooth}
	\end{equation}
\end{myassumption}
\begin{myassumption}{3a}\label{a:bounded-noise-opt}
	Let $f$ be convex, with global minimizer $x^\star$, and define
	\begin{align*}
		\zeta_n^2 & \coloneqq \normsq{\nabla f_n(x^\star)}_2, & \bar \zeta^2 & \coloneqq \textstyle \frac{1}{N}\sum_{n=1}^N \zeta_n^2.
	\end{align*}
	Similarly, define
	\begin{align*}
		\sigma_n^2 & \coloneqq \expec{\normsq{\nabla F_n(x^\star, \xi_n) - \nabla f_n(x^\star)}_2}, & \bar \sigma^2 & \coloneqq \textstyle \frac{1}{N}\sum_{n=1}^N \sigma_n^2.
	\end{align*}
	We assume that $\bar \sigma^2$ and $\bar \zeta^2$ are bounded.
\end{myassumption}
Here, $\bar \sigma^2$ measures the noise level and $\bar \zeta^2$ the heterogeneity, or diversity, of the functions $f_n$.
If all functions are identical, then $\bar \zeta^2 = 0$.
Prior to~\cite{unified_decentralized_SGD}, most work assumed bounded heterogeneity and noise \emph{everywhere}, whereas this assumption only requires bounds at $x^\star$.
\begin{theorem} \label{thm:convex}
	Given weight matrices (possibly extended with delays as in \cref{sec:incorporating-delays}) satisfying \cref{a:gossip}, cost functions satisfying \cref{a:bounded-noise-opt,a:lsmooth-stoch}, and a target accuracy $\epsilon > 0$, there exists a constant step-size $\eta$ such that \cref{alg:decentralized_opt_alg} reaches said accuracy $\frac{1}{K} \sum_{k=0}^{K-1} \mathbb{E} f(\tilde x^{[k]}) - f^\star \leq \epsilon$ after at most $K$ iterations, where $K$ is of order
	\begin{equation*}
		\mathcal{O} \left( \frac{\bar \sigma^2}{N \epsilon^2} + \frac{\beta\sqrt{L}(\bar \zeta \tau + \bar \sigma \sqrt{p\tau})}{p \epsilon^{3/2}} + \frac{\beta L \tau}{p\epsilon} \right) \cdot \normsq{\tilde x^{[0]} - x^*}_2.
	\end{equation*}
\end{theorem}
The proof of \cref{thm:convex} is provided in~\cref{ap:proof1}.

\subsection{Non-convex case}
For the non-convex case, we relax the $L$-smoothness assumption and only require it for the expectation of the local cost functions, i.e., $f_n$.
\begin{myassumption}{2b}\label{a:lsmooth}
	Each function $f_n(x)$ is $L$-smooth, that is, for all $x, y \in \R^d$:
	\begin{equation}
		\norm{\nabla f_n(x) - \nabla f_n(y) } \leq L \norm{x -y}. \label{eq:smooth_nc}
	\end{equation}
\end{myassumption}
\Cref{a:lsmooth} is more general than \cref{a:lsmooth-stoch}, since for convex $F_n(x, \xi_n)$, \cref{a:lsmooth-stoch} implies \cref{a:lsmooth}~\cite{Nesterov_2018}. 
Nevertheless, we use the stronger $L$-smoothness assumption in the convex case to allow the weak noise assumption of boundedness at $x^\star$.

For the non-convex case, where a unique $x^\star$ does not necessarily exist, we replace \cref{a:bounded-noise-opt} with the following assumption.
\begin{myassumption}{3b}\label{a:bounded-noise}
	We assume that there exists non-negative constants $P$ and $\hat \zeta$ such that for all $x \in \R^d$,
	\begin{align} \textstyle
		\frac{1}{N} \sum_{n = 1}^N \normsq{\nabla f_n(x)}_2 \leq \hat \zeta^2 + P \normsq{\nabla f(x)}_2, \label{eq:grad_opt_nc}
	\end{align}
	and constants  $M$ and $ \hat \sigma $ such that for all $x_1, \dots, x_N \in \R^d$,
	\begin{align} \textstyle
		\Psi \leq \hat \sigma^2 +  \frac{M}{N} \sum_{n = 1}^N \normsq{\nabla f_n(x_n)}_2, \label{eq:noise_opt_nc}
	\end{align}
	where $\Psi \coloneqq \frac{1}{N} \sum_{n = 1}^N \expec{\normsq{\nabla F_n(x_n, \xi_n) - \nabla f_n(x_n)}_2}$.
\end{myassumption}
These are standard assumptions from the literature. For example, further discussion on these assumptions and their comparisons with existing literature can be found in~\cite{unified_decentralized_SGD}.

\begin{theorem} \label{thm:non-convex}
	Given weight matrices (possibly extended with delays as in \cref{sec:incorporating-delays}) satisfying \cref{a:gossip}, cost functions satisfying \cref{a:bounded-noise,a:lsmooth}, and a target accuracy $\epsilon > 0$, there exists a constant step-size $\eta$ such that \cref{alg:decentralized_opt_alg} reaches said accuracy $\frac{1}{K}\sum_{k=0}^{K-1} \mathbb{E} \normsq{\nabla f(\tilde x^{[k]})}_2 \leq \epsilon$ after at most $K$ iterations, where $K$ is of order
	 \begin{equation*}
        \begin{aligned}
            & \mathcal{O} \left( \frac{ \hat \sigma^2 }{N \epsilon^2} + \frac{\beta(\hat \zeta \sqrt{M + 1} + \hat \sigma \sqrt{p\tau})}{p \epsilon^{3/2}} 
        \right) \cdot LF_0 \\
        & + \mathcal{O} \left( \frac{\beta\tau\sqrt{(P+1)(M+1)}}{p\epsilon} \right) \cdot LF_0.
        \end{aligned}
    \end{equation*}
\end{theorem}
The proof of \cref{thm:non-convex} is provided in~\cref{ap:proof2}.

\section{Experimental results} \label{sec:experimental-results}
\begin{figure*}[!t]
\centering
\subfloat[]{\includegraphics[width=3in]{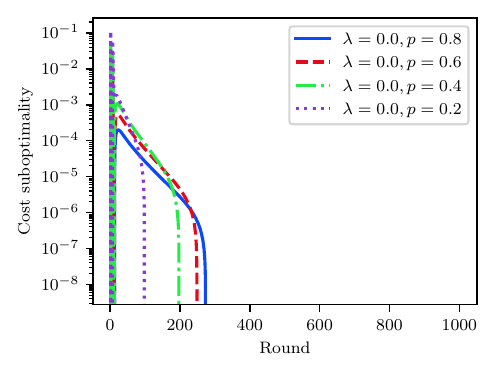}%
\label{fig:cost_p}}
\hfil
\subfloat[]{\includegraphics[width=3in]{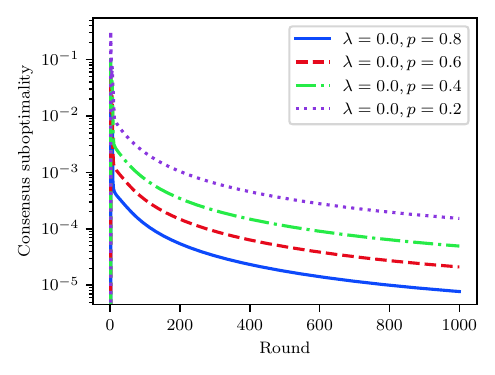}%
\label{fig:cons_p}}\\
\subfloat[]{\includegraphics[width=3in]{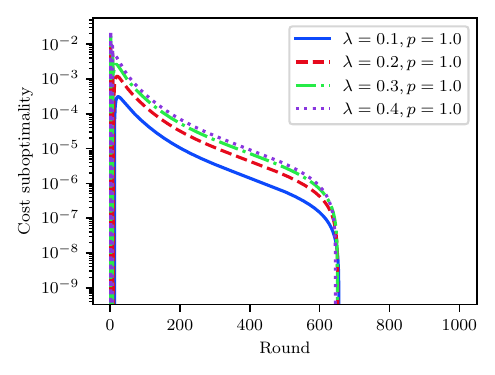}%
\label{fig:cost_lam}}
\hfil
\subfloat[]{\includegraphics[width=3in]{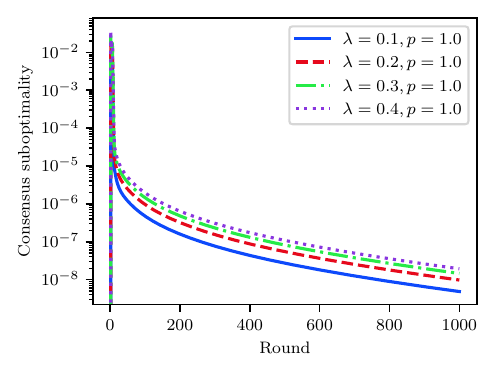}%
\label{fig:cons_lam}}\\
\subfloat[]{\includegraphics[width=3in]{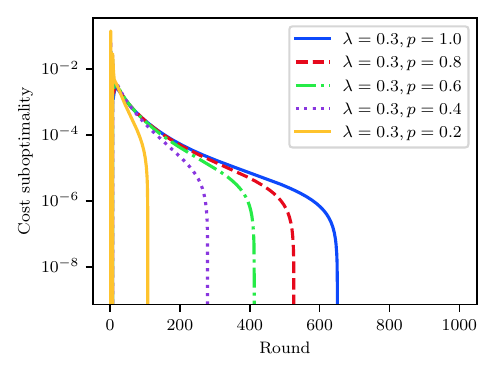}%
\label{fig:cost_mix}}
\hfil
\subfloat[]{\includegraphics[width=3in]{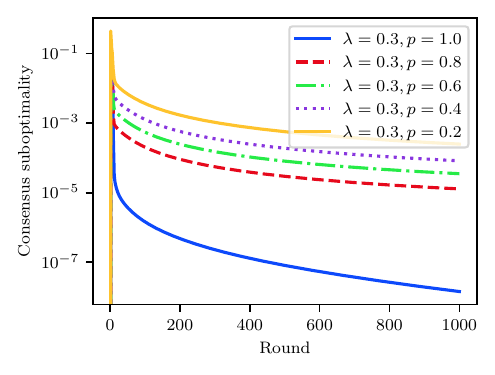}%
\label{fig:cons_mix}}
\caption{Cost and consensus suboptimality plots for various values of $\lambda$ and $p$. Cost is defined as $\frac{1}{N}\sum_{n=1}^N f_n(x_n^{[k]})$, and consensus is $\frac{1}{N}\sum_{n=1}^N (\bar x^{[k]} -x_n^{[k]})^2$. The suboptimality is the difference with respect to the baseline, $p=1$, which is centralized SGD. (a) Cost suboptimality plot for varying levels of $p$, without delays. (b) Consensus suboptimality plot for varying levels of $p$, without delays. (c) Cost suboptimality plot for varying levels of delays, with a complete graph. (d) Consensus suboptimality plot for varying levels of delays, with a complete graph. (e) Cost suboptimality for varying levels of $p$, with a fixed delay probability of $\lambda = 0.3$. (f) Consensus suboptimality for varying levels of $p$, with a fixed delay probability of $\lambda = 0.3$.}
\label{fig:regression_results}
\end{figure*}
In this section, we present experimental results conducted on a logistic regression problem with $\ell_2$ regularization, employing the \emph{mushrooms} dataset from LIBSVM \cite{LIBSVM}. 
This optimization problem, which is convex, consists of finding the weight vector $x$ that minimizes
\begin{equation}
f(x) = \frac{1}{N_s} \sum_{n=1}^{N_s} \log(1+\exp(-y_n x^\top s_n)) + \frac{\lambda}{2} \|x\|^2,
\end{equation}
where $N_s$ is the number of samples, $s_n$ is the $n$-th sample, $y_n$ is its corresponding label ($-1$ or $1$), and $\lambda$ is the regularization parameter.

We run simulations for 1,000 rounds with 1,024 warm-up rounds and 100 clients. 
The number of samples in the dataset is 8,124.
The learning rate and the $\ell_2$ regularization strength are set to 2 and $\tfrac{1}{8124}$, respectively.
For all experiments, each node uniformly averages all received messages and the weights are proportional to the inverse of the in-degree.

In what follows, we present results for time-invariant and time-varying communication networks in separate subsections.
The code for both cases is publicly available~\cite{code_for_dtgo}.

\subsection{Time-invariant graph experiments}

We categorize our time-invariant experiments into three distinct scenarios, each shedding light on different aspects of our approach.
For each scenario, we conduct 1,000 experiments (each running for 1,000 rounds and 1,024 warm-up rounds), generating a time-invariant graph for each experiment, and report the mean value of the results.

In the first scenario, we consider $G(N,p)$ Gilbert random graphs~\cite{gilbert}, where $N=100$ is the number of nodes and $p$ is a parameter that controls the sparsity of the graph, as each edge is included in the graph with probability $p$, independently from every other edge.
A higher value of $p$ implies a greater likelihood of edge existence and thus lower sparsity. 
Separable graphs are discarded and regenerated until they are strongly connected.
In the second scenario, we add delays to the full graph.
These delays follow a Poisson distribution with parameter $\lambda$, where a higher $\lambda$ signifies an increased average number of delays per link.
For the third scenario, we explore random graphs with varying $p$ values while introducing delays with a fixed parameter $\lambda=0.3$, to investigate the interplay between topology and delays.

It is worth noting that when $p=1$, the graph becomes complete.
Moreover, $\lambda=0$ implies no delays and for a complete graph without delays, our algorithm becomes centralized SGD, which serves as our benchmark.
Cost and consensus metrics are defined as 
\begin{align*}
    \frac{1}{N}\sum_{n=1}^N f_n(x_n^{[k]}) \quad \text{and} \quad \frac{1}{N}\sum_{n=1}^N (\bar x^{[k]} -x_n^{[k]})^2, 
\end{align*}
respectively. 
For both measures, the suboptimality is defined as the difference between the value and that of the centralized SGD method, which is equivalent to the case of $p=1$ and $\lambda=0$, as was previously mentioned.

\Cref{fig:cost_p,fig:cons_p} show how graph topology, quantified by the edge probability $p$, influences the performance of our algorithm. 
Higher graph connectivity accelerates consensus and a smaller consensus value results in solutions that diverge more at each node, consequently reducing local costs. 
This phenomenon is reflected in the faster disappearance of cost suboptimality for smaller values of $p$.

\Cref{fig:cost_lam,fig:cons_lam} display the effects of communication delays on our algorithm's performance.
As expected, higher values of $\lambda$ lead to slower convergence.

\Cref{fig:cost_mix,fig:cons_mix} confirm what was shown in both previous cases.
Namely, the presence of delays slows down consensus, but has a small effect on cost suboptimality.
Conversely, higher graph sparsity affects cost suboptimality more than the presence of delays, but does not hinder consensus as much.

\subsection{Time-varying graph experiments}
\begin{figure*}[!tbp]
\centering
\subfloat[]{\includegraphics[width=3in]{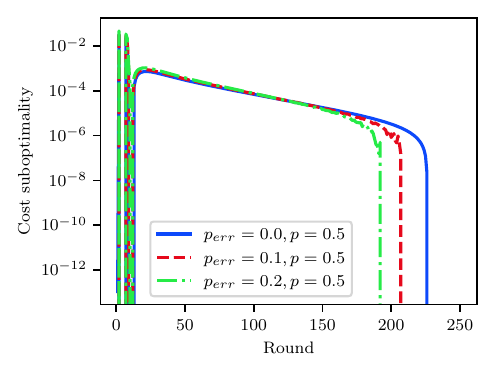}%
\label{fig:cost_q}}
\hfil
\subfloat[]{\includegraphics[width=3in]{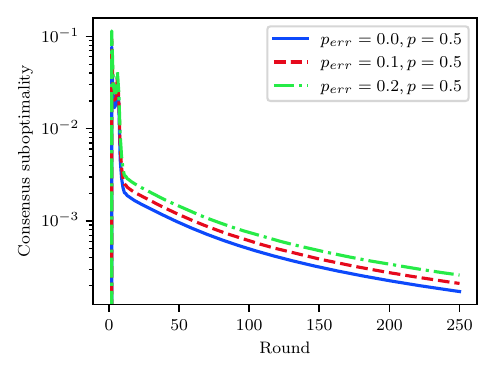}%
\label{fig:cons_q}}
\caption{Cost and consensus suboptimality plots for various values of $p_{err}$. Cost is defined as $\frac{1}{N}\sum_{n=1}^N f_n(x_n^{[k]})$ and consensus is $\frac{1}{N}\sum_{n=1}^N (\bar x^{[k]} -x_n^{[k]})^2$. The suboptimality is the difference with respect to the baseline, $p=1$, which is centralized SGD. (a) Cost suboptimality plot for different values of $p_{err}$. (b) Consensus suboptimality plot for different values of $p_{err}$.}
\label{fig:time_varying_regression_results}
\end{figure*}
For our time-varying experiments, we simulate a typical situation found in real-world networks, where the topology changes from round to round due to links drop at certain rounds.
Each experiment begins with a randomly generated $G(100, 1/2)$ graph $G_0$, including all possible self-loops.
At each round, we consider that an edge between two distinct nodes has an error probability $p_{err}$ and in the case of an error disappears from the graph for that round.

To account for the larger variance due to the time-varying topology, we present results averaged over 40,000 experiments, each of 250 rounds.

\Cref{fig:time_varying_regression_results} displays the simulation results, which are consistent with the findings from \cref{fig:regression_results}.
Observe that higher error probabilities affect consensus negatively.
Thus, as nodes drift further away from the consensus, they can obtain lower local costs, which is reflected in the cost plot.

\section{Conclusions} \label{sec:conclusions}
We have addressed the problem of decentralized optimization in networks with communication delays and directed communication graphs.
The key contribution of our work is the introduction of a novel gossip-based optimization algorithm, named \algname, which circumvents the requirement for knowledge of out-degrees.
This algorithm opens new avenues for decentralized optimization in various real-world scenarios, such as networks with delays in their communication links or limited acknowledgment capabilities.

We have studied the performance of \algname through theoretical analysis and numerical simulations. 
Our analysis includes time-varying directed communication networks.
Importantly, we have established convergence guarantees for both convex and non-convex objectives under mild assumptions. 
Our analysis shows that the effects of topology and delays are confined to higher-order terms, and the main error term remains the same as that of plain centralized SGD.

Furthermore, our algorithm offers practical advantages, including its ability to operate without the need for multiple rounds of gossip per local optimization step or bounded gradient assumptions.
By generalizing decentralized Stochastic Gradient Descent, our algorithm can be applied in bidirectional communication graphs, ensuring compatibility with existing decentralized optimization frameworks.

In summary, our work presents a novel decentralized optimization approach, that addresses the challenges posed by communication delays and time-varying directed network topologies. 
We believe that this work can lead to applications in diverse fields such as collaborative machine learning, sensor networks, and multi-agent systems.

\begin{appendices}
\section{Lemmas and remarks}
For our proof, we generalize the analysis framework for convex and non-convex decentralized optimization from~\cite{unified_decentralized_SGD} to non-doubly stochastic matrices and our re-weighted optimization iterations in \algname, as we require.
The following results are needed to prove \cref{thm:convex,thm:non-convex}.

\begin{lemma}[Descent recursion for the convex case]\label{l:descent-convex}
    Under \cref{a:bounded-noise-opt,a:lsmooth-stoch,a:gossip}, with $\eta \leq \frac{1}{12L}$, we can ensure
	\begin{multline}
		\expec{\normsq{\tilde x^{[k+1]}-x^\star}} \leq \normsq{\tilde x^{[k]} - x^\star} + \frac{\eta^2 \bar \sigma^2}{N}  \\
		{}- \eta (f(\tilde x^{[k]}) - f(x^\star))  + \eta \frac{3L}{N} \sum_{n=1}^N \normsq{x_n^{[k]} - \tilde x^{[k]}}.
	\end{multline}
\end{lemma}
\begin{IEEEproof}
    Using the matrix form of \algname iterations from \cref{eq:alg-iterations}, the weighted average at Round $k$ is
    \begin{equation*}
        \tilde X^{[k+1]} = W^\infty W^{[k]}\left(X^{[k]} - \eta D \partial F\left(X^{[k]}\right) \right).
    \end{equation*}
    Observe that $W^\infty D = J$, where $J$ is the average matrix whose entries are all $1/N$.
    Then, using $W^\infty W^{[k]} = W^\infty$ from Item (i) in \cref{a:gossip}, we obtain
    \begin{equation} \label{eq:matrix-descent}
        \tilde X^{[k+1]} = \tilde X^{[k]} - \eta J \partial F\left(X^{[k]}\right).
    \end{equation}
    In vector notation, \cref{eq:matrix-descent} is equivalent to
    \begin{equation} \label{eq:vector-descent}
        \tilde x^{[k+1]} = \tilde x^{[k]} - \frac{\eta}{N} \sum_{n=1}^N \nabla F_n(x_n^{[k]}, \xi_n).
    \end{equation}
    Note that this implies that our algorithm preserves the \emph{weighted} average of previous iterations, plus the un-weighted average of the gradient steps at each node.
    Using this observation, the rest of the proof follows as in~\cite[Lemma 8]{unified_decentralized_SGD} and is omitted in this manuscript for brevity.
\end{IEEEproof}

\begin{lemma}[Descent recursion for the non-convex case]\label{l:descent-non-convex}
    Under \cref{a:bounded-noise,a:lsmooth,a:gossip}, with $\eta \leq \frac{1}{4L(M+1)}$, we can ensure
	\begin{multline}
		\expec{f(\tilde x^{[k+1]})} \leq f(\tilde x^{[k]}) + \frac{\eta^2 L \hat \sigma^2}{N}  \\
		{}- \frac{\eta}{4}\normsq{\nabla f(\tilde x^{[k]})} + \eta \frac{L^2}{N} \sum_{n=1}^N \normsq{x_n^{[k]} - \tilde x^{[k]}}.
	\end{multline}
\end{lemma}
\begin{IEEEproof}
    Using the same preservation of the weighted average argument from~\cref{eq:vector-descent}, the proof follows as in~\cite[Lemma 10]{unified_decentralized_SGD} and is omitted for brevity.
\end{IEEEproof}

\begin{lemma}[Consensus distance recursion for the convex case]\label{l:recursion-convex}
	Let $g^{[k]} = \frac{1}{N} \sum_{n=1}^N  \mathbb{E} \normsq{x_n^{[k]} - \tilde x^{[k]}}_2$. Under \cref{a:bounded-noise-opt,a:lsmooth-stoch,a:gossip}, and $\eta \leq \frac{p}{12 \tau L \beta }$, we can ensure
	\begin{multline}
		g^{[k]} \leq \left(1 - \frac{p}{2} \right) g^{[k-\tau]} + \frac{p}{16\tau} \sum_{\ell=k-\tau}^{k-1} g^{[\ell]} \\
		{} +  \frac{ 18 L \beta^2 \tau}{p} \sum_{\ell=k-\tau}^{k-1} \eta^2 (f(\tilde x^{[\ell]}) - f (x^\star))
		\\ + \beta^2(\bar \sigma^2 + \frac{9\tau}{p} \bar \zeta^2)  \sum_{\ell=k-\tau}^{k-1} \eta^2.
	\end{multline}
\end{lemma}
\begin{IEEEproof}
    By definition of $g^{[k]}$,
    \begin{equation} \label{eq:start-of-gossip}
        N g^{[k]} = \mathbb{E} \normsq{X^{[k]} - \tilde X^{[k]}}_F.
    \end{equation}
    \algname iterations in matrix notation satisfy \cref{eq:alg-iterations} and $\tilde X^{[k]} = W^\infty X^{[k]}$. 
    Therefore, defining $T\coloneqq X^{[k]} - \tilde X^{[k]}$,
    \begin{equation}
     T =(W^{[k]} - W^\infty) \left(X^{[k-1]} - \eta D \partial F\left(X^{[k-1]}\right) \right),
    \end{equation}
    where we have used  $W^\infty W^{[k]} = W^\infty$ from Item (i) in~\cref{a:gossip}.
    Note that $(W^{[k]} - W^\infty)\tilde X^{[k-1]} = 0$, by the same argument.
    Thus, we obtain
    \begin{align*}
    T & =(W^{[k]} - W^\infty) \left(X^{[k-1]} - \tilde X^{[k-1]} \right) \\
    & - \eta (W^{[k]} - W^\infty)   D \partial F\left(X^{[k-1]}\right).
    \end{align*}
    We repeat this procedure $\tau$ times and the term $T$ becomes
    \begin{align*}
        T & = \left( \prod_{\ell = k-\tau}^{k-1}(W^{[\ell]} - W^\infty) \right) (X^{[k-\tau]} - \tilde X^{[k-\tau]} )\\
        & - \eta \sum_{\ell = k - \tau}^{k-1} \left(\prod_{m=\ell}^{k-1} (W^{[m]} - W^\infty) \right) D \partial F\left(X^{[\ell]}\right) .
    \end{align*}
    Note that for $k < \tau$, we can simply define the gossip matrices as the identity, and the gradients as zero, and the expression holds.
    Now, we simplify the products using $W^\infty(W^{[k]} - W^\infty) = 0$, for all $k$, and obtain
    \begin{align}
        T &= \underbrace{\left(\left( \prod_{\ell = k-\tau}^{k-1}W^{[\ell]} \right) - W^\infty \right) X^{[k-\tau]}}_{T_1} \nonumber \\
        & - \eta \underbrace{\sum_{\ell = k - \tau}^{k-1} \left( \left(\prod_{m=\ell}^{k-1} W^{[m]}\right) D - J \right)  \partial F\left(X^{[\ell]}\right)}_{T_2} .
    \end{align}
    Finally, we plug this into~\cref{eq:start-of-gossip} and observe that the term $T_1$ can be bounded using Item (iii) in \cref{a:gossip} and the sum of gossip matrix products can be bounded using Item (iv) in \cref{a:gossip}.
    The rest of the proof follows as in~\cite[Lemma 9]{unified_decentralized_SGD} and is omitted for brevity.
\end{IEEEproof}

\begin{lemma}[Consensus distance recursion for the non-convex case]\label{l:recursion-non-convex}
	Let $g^{[k]} = \frac{1}{N} \sum_{n=1}^N  \mathbb{E} \normsq{x_n^{[k]} - \tilde x^{[k]}}_2$. Under \cref{a:bounded-noise,a:lsmooth,a:gossip}, and $\eta \leq \frac{p}{12 \tau L \beta }$, we can ensure
	\begin{multline}
		g^{[k]} \leq \left(1 - \frac{p}{2} \right) g^{[k-\tau]} + \frac{p}{16\tau} \sum_{\ell=k-\tau}^{k-1} g^{[\ell]} \\
		{} +  2P\beta^2 (\frac{3\tau}{p} + M) \sum_{\ell=k-\tau}^{k-1} \eta^2 \normsq{\nabla f (\tilde x ^{[\ell]})}
		\\ + 2\beta^2 (\hat \sigma^2 + (\frac{3\tau}{p} + M) \hat \zeta^2)  \sum_{\ell=k-\tau}^{k-1} \eta^2.
	\end{multline}
\end{lemma}
\begin{IEEEproof}
    We follow the derivations of \cref{l:recursion-convex}, presented above, and conclude as in the proof for~\cite[Lemma 11]{unified_decentralized_SGD}, which is omitted for brevity.
\end{IEEEproof}

\begin{lemma}\label{l:order}{\normalfont (Re-stated from~\cite[Lemma 14]{unified_decentralized_SGD})}
    If non-negative sequences $e^{[0]}, \ldots e^{[K]}$ and $r^{[0]}, \ldots r^{[K]}$ satisfy
    \begin{equation} \label{eq:result}
    		\frac{1}{2K} \sum_{k=0}^{K-1}be^{[k]} \leq \frac{r^{[0]}}{T\eta} + c \eta + 64AB \eta^2,
    \end{equation}
    where $b>0$, $c,A,B\geq 0$, then there exists a constant positive step-size $\eta < \frac{1}{d}$ such that
    \begin{equation*}
        \frac{1}{K}\sum_{k=0}^{K-1} e^{[k]} \leq \mathcal{O} \left( \left( \frac{c r^{[0]}}{K}\right)^{\frac{1}{2}} + (BA)^{\frac{1}{3}}\left( \frac{r^{[0]}}{K}\right)^{\frac{2}{3}} + \frac{d r^{[0]}}{K} \right).
    \end{equation*}
\end{lemma}

\section{Proof of Theorem \ref{thm:convex}} \label{ap:proof1}
    The main goal of the proof is to derive an upper bound on the distance to the solution, which is $r^{[k]} = \mathbb{E} \normsq{\tilde x^{[k]} - x^\star}$.
    Let us start by using \cref{l:descent-convex} to obtain
    \begin{equation} \label{eq:dist-to-sol}
    	r^{[k+1]} \leq r^{[k]} - b \eta e^{[k]} + c \eta^2 + \eta B g^{[k]},
    \end{equation}
    where $g^{[k]} = \frac{1}{N} \sum_{n=1}^N \mathbb{E}\normsq{\tilde x^{[k]} - x_n^{[k]}}$, $e^{[k]} = f(\tilde x^{[k]}) - f(x^\star)$, $b=1$, $c = \frac{\bar \sigma^2}{N}$, and $B = 3L$.
    Note that $g^{[k]}$ is a measure of the mean distance to the weighted average of models and corresponds to the equal-weight average in the doubly stochastic case, since all weights would be $1/N$, as previously discussed.
    
    Now, we bound the distance to the weighted average, or consensus distance, with the recursion obtained from \cref{l:recursion-convex}:
    \begin{equation} \label{eq:dist-to-avg}
	   \begin{aligned}
    		g^{[k]} & \leq \left( 1 - \frac{p}{2} \right) g^{[k-\tau]} + \frac{p}{16\tau} \sum_{\ell=k-\tau}^{k-1} g^{[\ell]} \\
    		    & + D' \sum_{\ell=k-\tau}^{k-1}\eta^2 e_\ell + A  \sum_{\ell=k-\tau}^{k-1} \eta^2,
    	\end{aligned}
    \end{equation}
    where $A = \beta^2(\bar \sigma^2 + \frac{9\tau}{p} \bar \zeta^2)$ and $D'= \frac{ 18 L \beta \tau}{p}$.
    Note that for $k\leq \tau$, we can define $g^{[k]} = 0$, $e^{[k]} = 0$, and the recursion in \cref{eq:dist-to-avg} holds.
Next, for any positive integer $K$, we add \cref{eq:dist-to-avg} from $k=0$ to $K-1$ and simplify to obtain
\begin{equation}
    \begin{aligned}
        \sum_{k=0}^{K-1} g^{[k]} &\leq \left( 1 - \frac{p}{4} \right) \sum_{k=0}^{K-1} g^{[k]} + D' \sum_{k=0}^{K-1} \tau \eta^2 e^{[k]} \\
        &+ A  T \tau \eta^2.
    \end{aligned}
\end{equation}
We multiply both sides by $B$ and re-arrange, resulting in
\begin{equation} \label{eq:simplified-recursion}
	B \sum_{k=0}^{K-1}g^{[k]} \leq \frac{b}{2} \sum_{k=0}^{K-1} e^{[k]} + 64 BA \frac{\tau}{p}K\eta^2,
\end{equation}
as long as $\eta \leq \frac{1}{16}\sqrt{\frac{pb}{D'B\tau}}$.
Then, we substitute~\cref{eq:simplified-recursion} into the last term of~\cref{eq:dist-to-sol}, added from $0$ to $K-1$, and divided by $\eta$ to obtain
$$\frac{1}{2} \sum_{k=0}^{K-1}be^{[k]} \leq \sum_{k=0}^{K-1}\frac{r^{[k]} - r^{[k+1]}}{\eta} + Kc \eta + K64AB \eta^2.$$
We divide by $K$ and telescope the sum, resulting in
\begin{equation} \label{eq:final-result}
	\begin{aligned}
		\frac{1}{2K} \sum_{k=0}^{K-1}be^{[k]} \leq \frac{r^{[0]} - r^{[K]}}{K\eta} + c \eta + 64AB \eta^2.
	\end{aligned}
\end{equation}
We conclude by applying~\cref{l:order} to \cref{eq:final-result}, with $r^{[k]} = \mathbb{E} \normsq{\tilde x^{[k]} - x^\star}$, $e^{[k]} = f(\tilde x^{[k]}) - f(x^\star)$, $b=1$, $c = \frac{\bar \sigma^2}{N}$, $d = \frac{p}{12\tau L\beta}$, $A = \beta^2(\bar \sigma^2 + \frac{9\tau}{p} \bar \zeta^2)$, and $B = 3L$.

\section{Proof of Theorem \ref{thm:non-convex}} \label{ap:proof2}
    The proof follows the same line as the convex case, but now the distance to the solution is $r^{[k]} = \mathbb{E} f(\tilde x^{[k]}) - f^\star$, since we do not have a global minimum.
    Using \cref{l:descent-non-convex}, we arrive to the expression in \cref{eq:dist-to-sol}, only now $r^{[k]} = \mathbb{E} f(\tilde x^{[k]}) - f^\star$, $e^{[k]} = \normsq{\nabla f(\tilde x^{[k]})}_2$, $b = \frac{1}{4}$, $c = \frac{L \hat \sigma^2}{N}$, and $B = L^2$.
    We can obtain a consensus distance recursion, similar to that of \cref{eq:dist-to-avg} with \cref{l:descent-non-convex}, where $A = 2\beta^2 (\hat \sigma^2 + (\frac{3\tau}{p} + M) \hat \zeta^2)$ and $D' = 2P\beta^2 (\frac{3\tau}{p} + M)$.
    The rest of the proof follows unmodified, and we obtain the theorem statement by applying \cref{l:order} to \cref{eq:final-result} with the new constants, where $d = 32 L\beta \sqrt{2 (P+1) (\frac{3\tau}{p} + M) \frac{\tau}{p}}$.

\end{appendices}

\bibliography{references}
\bibliographystyle{IEEEtran}

\end{document}